\documentclass[11pt]{article}
\usepackage{graphicx}
\usepackage{coling2020}
\usepackage{times}
\usepackage{url}
\usepackage{latexsym}
\usepackage{array}
\usepackage{multirow}
\usepackage{multicol}
\usepackage{caption}
\usepackage{subcaption}
\usepackage{latexsym}
\usepackage{amstext}
\usepackage{xspace}
\usepackage{bm}
\usepackage{color}
\usepackage{xcolor}
\newcommand{\paratitle}[1]{\vspace{1.5ex}\noindent\textbf{#1}}

\newcommand{\ie}{\emph{i.e.,}\xspace}

\newcommand{\eg}{\emph{e.g.,}\xspace}
\newcommand{\ignore}[1]{}
\colingfinalcopy % Uncomment this line for the final submission

\title{Towards Topic-Guided Conversational Recommender System}

\author{Kun Zhou$^{1\dagger}$,
 Yuanhang Zhou$^{1\dagger}$,
 Wayne Xin Zhao$^{2,3}$\footnotemark[1],
 Xiaoke Wang$^1$ and 
 Ji-Rong Wen$^{2,3}$
 \\
 $^1$School of Information, Renmin University of China\\
 $^2$Gaoling School of Artificial Intelligence, Renmin University of China\\
 $^3$Beijing Key Laboratory of Big Data Management and Analysis Methods\\
 \tt{francis\_kun\_zhou@163.com},
 \tt{\{sdzyh002,batmanfly\}@gmail.com}\\
 \tt{\{yarkonawang,jrwen\}@ruc.edu.cn},
}

\date{}

\begin{document}
\maketitle

\renewcommand{\thefootnote}{\fnsymbol{footnote}}
% \footnotetext[1]{Equal contribution.}
\footnotetext[1]{Corresponding author. $\dagger$Equal contribution}

\begin{abstract}
Conversational recommender systems (CRS) aim to recommend high-quality items to users through interactive conversations.
To develop an effective CRS, the support of high-quality datasets is essential.
Existing CRS datasets mainly focus on immediate requests from users, while lack proactive guidance to the recommendation scenario. 
In this paper, we contribute a new CRS dataset named \textbf{TG-ReDial} (\textbf{Re}commendation through \textbf{T}opic-\textbf{G}uided \textbf{Dial}og). Our dataset has two major features. First, it incorporates topic threads to enforce natural semantic transitions towards the recommendation scenario. Second, it is created in a semi-automatic way, hence human annotation is more reasonable and controllable.
Based on TG-ReDial, we present the task of topic-guided conversational recommendation, and propose an effective approach to this task.
Extensive experiments have demonstrated the effectiveness of our approach on three sub-tasks, namely topic prediction, item recommendation and response generation.
TG-ReDial is available at \textcolor{blue}{\url{https://github.com/RUCAIBox/TG-ReDial}}.

\ignore{only focused on finishing the recommendation task quickly, but neglect the topic guiding process in conversation and the persuasive reasons for recommendation.
To address these issues, we present a new dataset named \textbf{Re}commendation through \textbf{T}opic-\textbf{G}uided \textbf{Dial}og (\textbf{TG-ReDial}, which is built by human annotation based on real-world data.
Besides the rich background information about users, each dialogue in TG-ReDial is affiliated with a topic thread, while each recommendation in dialogue is accompanied with a persuasive reason.
Based on TG-ReDial, we present a novel task, topic-guided conversational recommender system (TG-CRS) consisting of three sub-tasks, and propose a framework to handle them.
Extensive experiments have demonstrated the effectiveness of our framework in yielding better performance on TG-CRS task.
}
\end{abstract}

\section{Introduction}
Recently, conversational recommender system (CRS)~\cite{DBLP:conf/emnlp/ChenLZDCYT19,DBLP:conf/sigir/SunZ18,DBLP:conf/nips/LiKSMCP18,DBLP:conf/cikm/ZhangCA0C18,DBLP:journals/corr/abs-1907-00710} has become an emerging research topic, which aims to provide high-quality recommendations to users through natural language conversations. 
Generally, a CRS is composed of a recommender component and a dialog component, which make suitable recommendation and generate proper response, respectively.
To develop an effective CRS,  high-quality datasets are crucial to learn the model parameters. Existing CRS datasets roughly fall into two main categories, namely attribute-based user simulation~\cite{DBLP:conf/sigir/SunZ18,DBLP:conf/wsdm/Lei0MWHKC20,DBLP:conf/cikm/ZhangCA0C18} and 
chit-chat based goal completion~\cite{DBLP:conf/nips/LiKSMCP18,DBLP:conf/emnlp/ChenLZDCYT19,DBLP:conf/acl/LiuWNWCL20}. 
% The recommender component is to infer user`s preferred items using the historical utterances in conversation and the personal information of the user, while the dialog component aims to generate proper response with the purpose of obtaining preference information or giving persuasive recommendation. 

%Similar to salesman in real world, CRS focuses on how to proactively clarify user`s preference and provides persuasive recommendation through dialog-based interaction. Therefore, CRS is expected to guide the user and achieve accurate recommendation in E-commerce platform.

%In previous research~\cite{DBLP:journals/corr/abs-1907-00710,DBLP:conf/emnlp/ChenLZDCYT19,DBLP:conf/sigir/SunZ18,DBLP:conf/nips/LiKSMCP18}, CRS is generally composed by a recommender component and a dialog component. The recommender component is to infer user`s preferred items using the historical utterances in conversation and the personal information of the user, while the dialog component aims to generate proper response with the purpose of obtaining preference information or giving persuasive recommendation. 

\begin{figure}[tb] %图片浮动环境，类似表格中的 table [htp] 参数和表格的类似
  %\vspace{-2mm}
  \centering %图片居中
  \includegraphics[width=\textwidth]{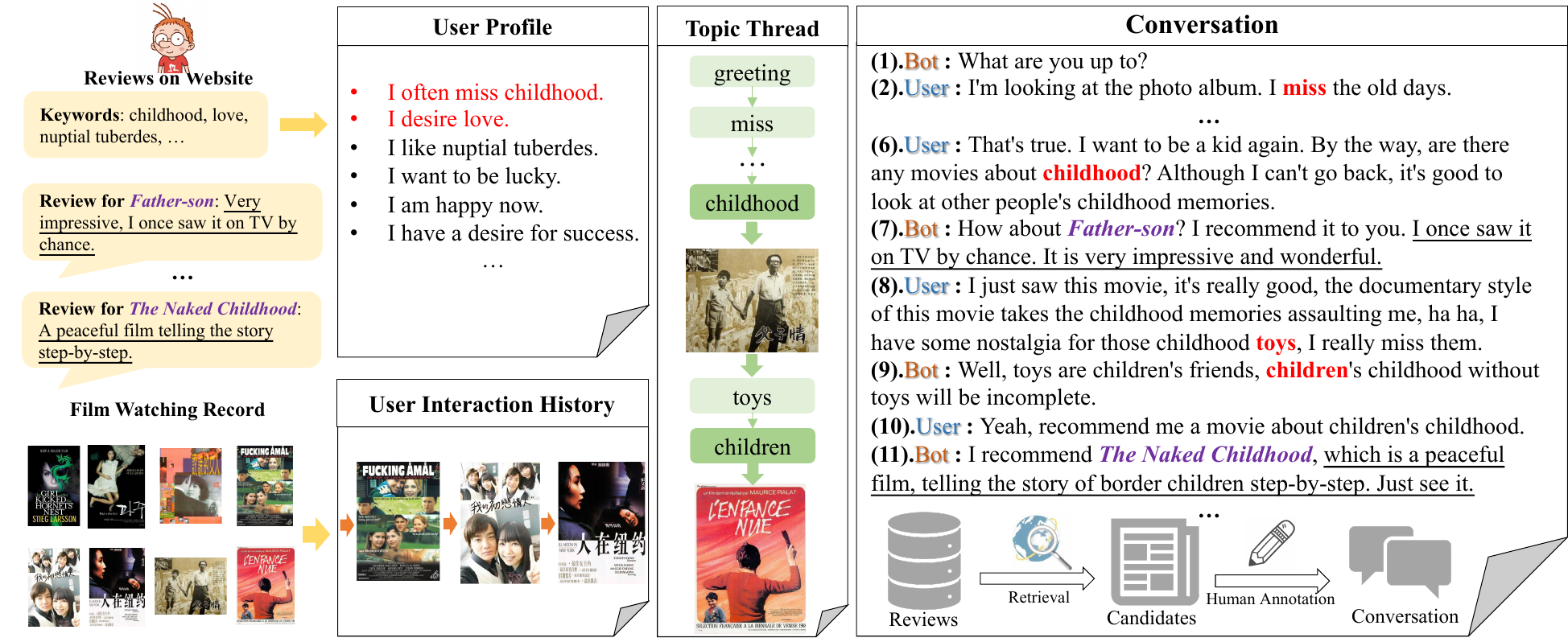}
  \caption{An illustrative example for TG-ReDial dataset. We utilize real data to construct the recommended movies, topic threads, user profiles and utterances.  Other user-related information (\eg historical interaction records) is also available in our dataset.
  %The whole dialog derives from real-world movie viewing record, and is grounded on topic thread, which is built considering the user profile. Each goal specifies a dialog type and a dialog topic (an entity). In conversation, we use different colors to indicate topic words (red) and recommended movies (purple), and use underline to indicate persuasive reasons from recommendation.
  }
  \label{fig-sample}
\end{figure}

These datasets usually assume that a user has clear, immediate requests when interacting with the system. They lack the proactive guidance (or transitions) from non-recommendation scenarios to the desired recommendation scenario.  
Indeed, it has become increasingly important that recommendations can be naturally triggered according to conversation context~\cite{DBLP:conf/acl/TangZXLXH19,DBLP:conf/emnlp/KangBSCBW19}. This issue has been explored to some extent by DuRecDial dataset~\cite{DBLP:conf/acl/LiuWNWCL20}. DuRecDial has characterized the goal-planning process by constructing a goal sequence. However, it mainly focuses on type switch or coverage for dialog sub-tasks (\eg non-recommendation, recommendation and question-answering).  Explicit semantic transition that leads up to the recommendation has not been well studied or discussed in DuRecDial dataset.
Besides, most of existing CRS datasets~\cite{DBLP:conf/acl/LiuWNWCL20,DBLP:conf/nips/LiKSMCP18} mainly rely on human annotators to create user profiles or generate the conversations. It is difficult to capture rich, complicated cases from real-world applications with a limited number of human annotators, since the generated conversations mainly reflect the characteristics (\eg interest) of annotators or predefined identities.
To tackle the above problems, we construct a new CRS dataset named \textbf{Re}commendation through \textbf{T}opic-\textbf{G}uided \textbf{Dial}og (\textbf{TG-ReDial}). 
It consists of 10,000 two-party dialogues between a seeker and a recommender in the movie domain. There are two new features in our dataset. 
First, we explicitly create a topic thread to guide the entire content flow for each conversation. Starting with a non-recommendation topic, the topic thread naturally guides the user to the recommendation scenario through a sequence of evolving topics. Our dataset enforces natural transitions towards recommendation through chit-chat conversations. Second, our dataset has been created in a semi-automatic way by involving reasonable and controllable human annotation efforts. The key idea is to align user identities in conversations with real users from a popular movie review website. In this way, the recommended movies, the created topic threads and the recommendation reasons are mined or generated based on real-world data. The major role of the human annotators is to revise, polish or rewrite the conversation data when necessary. Therefore, we do not rely on human annotators to create personalized user profiles as previous studies~\cite{DBLP:conf/nips/LiKSMCP18,DBLP:conf/acl/LiuWNWCL20}, making our conversation data closely resembles real-world cases. %Besides, we can also obtain historical interaction data for the linked user identities from the review website. 
Figure~\ref{fig-sample} presents an illustrative example for our TG-ReDial dataset.

%The core topic threads are mined and generated using real-world movie review dataset. 

\ignore{
To build TG-ReDial, we utilize the user`s real movie viewing record and generate topic thread for each conversation considering the user`s feedback.
Then we provide high-quality templates and ask annotators to complete the conversation based on the given prompts.
Hence the dialogues in TG-ReDial are not only informative and natural, but also consistent with the topic-guided conversational recommendation scenario.
}
%each dialogue is generated by rewriting and extending the user`s movie viewing record from real world.

\ignore{
We illustrate an example in Figure~\ref{fig-sample}.
In TG-ReDial, every dialogue owns a topic thread, in which the topics are gradually close to the target topic considering user`s feedback.
%When the conversation starts, the bot should proactively and naturally lead the conversation into a target topic. It can be viewed as a salesman who induces the user to buy highly profitable goods.
After reach the target topic, the bot should recommend user preferred item with a persuasive reason, in which the items derive from the real data and the reason is provided by human annotated comments.
%Furthermore, we consider recommending multiple items in each dialogue so that the bot needs to lead the conversation jumping among different items. It is a simulation of the scene that a salesman follows the customer around in the shop and help her/him find more desired goods.
Moreover, TG-ReDial provides rich background information such as user interaction history (item sequence) and user profile (preferred topic), which are helpful to CRS but usually lacked in other CRS datasets.
Therefore, we can utilize some advanced sequential recommendation models~\cite{DBLP:conf/icdm/KangM18,DBLP:conf/icdm/LiuWWLW16} and text understanding models~\cite{DBLP:conf/naacl/DevlinCLT19} to further improve CRS.
}

Based on the TG-ReDial dataset, we study a new task of topic-guided conversational recommendation, which can be decomposed into three sub-tasks, namely \emph{item recommendation}, \emph{topic prediction}, and \emph{response generation}. % Given the user profile and interaction history, it is mainly decomposed into three sub-tasks, namely \emph{item recommendation}, \emph{topic prediction}, and \emph{response generation}. 
Topic prediction aims to create the topic thread that leads to the final recommendation; item recommendation provides suitable items that meet the user needs; and response generation produces proper reply in natural language. In our approach, the recommender module utilizes both historical interaction and dialog text for deriving accurate user preference, which are modeled by sequential recommendation model SASRec~\cite{DBLP:conf/icdm/KangM18} and pre-trained language model BERT~\cite{DBLP:conf/naacl/DevlinCLT19}, respectively. 
The dialog module consists of a topic prediction model and a response generation model.
The topic prediction model integrates three kinds of useful data (\ie historical utterances, historical topics and user profile) to predict the next topic.
The response generation model is implemented based on  GPT-2~\cite{radford2019language} to produce responses for guiding users or giving persuasive recommendation.
To validate the effectiveness of our approach, we conduct extensive experiments on TG-ReDial dataset to compare our approach with competitive baseline models.

Our main contributions are summarized as follows:

(1) We release a new dataset TG-ReDial for conversational recommender systems. It emphasizes natural topic transitions that leads to the final recommendation. Our dataset is created in a semi-automatic way, and hence human annotation is more reasonable and controllable.
%in that all user identities are from a popular movie review website, and human annotation has been reas. 
%, which reduces the gap between CRS datasets and real-world application by proactively guide the conversation with the mined topic threads.

%(2) We incorporate rich background information (\emph{e.g.} user interaction history) into TG-CRS, so that we can utilize some advanced techniques in other domains to improve CRS.

(2) Based on TG-ReDial, we present the task of topic-guided conversational recommendation, consisting of item recommendation, topic prediction and response generation. We further develop an effective solution  to leverage multiple kinds of data signals based on Transformer and its variants BERT and GPT-2. 
%Furthermore, we design an effective approach to leveraging multiple kinds of data signals for conversational recommendation.

%We identify the task of topic-guided conversation recommender system (TG-CRS) containing three sub-tasks: \emph{item recommendation}, \emph{topic prediction}, and \emph{response generation}. We propose a framework consisting of a recommendation and a dialog module to handle them.

%(3) Extensive experiments conducted on TG-ReDial demonstrate the effectiveness of our proposed framework. 

\section{Related Work}
%In this section, we present the related works in three aspects, namely conversational  systems, conversational recommender systems and dataset for conversational recommendation.

\subsection{Conversation System}
Conversation systems~\cite{DBLP:conf/acl/ShangLL15,DBLP:conf/naacl/LiGBGD16,DBLP:conf/acl/DhingraLLGCAD17} study how to generate proper responses given multi-turn contextual utterances. Existing works can be categorized into task-oriented systems~\cite{DBLP:conf/acl/DhingraLLGCAD17,DBLP:conf/icassp/YoungSWY07} to accomplish specific goals (\eg book the ticket) and chit-chat systems~\cite{DBLP:conf/naacl/LiGBGD16,DBLP:conf/acl/ShangLL15,DBLP:conf/emnlp/ZhouZWLY19} to provide general-purpose dialogue.
Related to our work, topical information  have attracted much research interests in the research community for conversation systems~\cite{DBLP:conf/aaai/XingWWLHZM17,DBLP:conf/acl/TangZXLXH19,DBLP:conf/acl/XuWNWCL20}, since it can enhance the semantics of the generated conversation.
Early works~\cite{DBLP:conf/aaai/XingWWLHZM17,DBLP:conf/ijcai/LianXWPW19} focused on guiding the conversation topic for the next response, while recent studies~\cite{DBLP:conf/acl/TangZXLXH19,DBLP:conf/acl/XuWNWCL20} started to emphasize the multi-turn topic-guided process in the whole conversation. For example,
keyword transition~\cite{DBLP:conf/acl/TangZXLXH19} and knowledge graph~\cite{DBLP:conf/acl/XuWNWCL20} are incorporated to improve the topic-guided conversation systems.

\subsection{Conversational Recommender System}
Conversational recommender system (CRS)~\cite{DBLP:conf/emnlp/ChenLZDCYT19,DBLP:conf/sigir/SunZ18,DBLP:conf/nips/LiKSMCP18} aims to provide high-quality recommendation through conversations with users.
Generally, it consists of a dialog component to interact with a user and a recommender component to select items for recommendation considering user preference.
Early conversational recommender systems~\cite{DBLP:conf/kdd/Christakopoulou16,DBLP:conf/sigir/SunZ18,DBLP:conf/cikm/ZhouZWWZWW20} mainly asked questions about user preference over pre-defined slots to make recommendations.
Recently, several studies~\cite{DBLP:conf/nips/LiKSMCP18,DBLP:conf/emnlp/ChenLZDCYT19,DBLP:conf/acl/LiuWNWCL20} started to interact with user through natural language conversation, emphasizing fluent response generation and precise recommendation.
Furthermore, follow-up studies~\cite{DBLP:conf/emnlp/ChenLZDCYT19,DBLP:conf/emnlp/KangBSCBW19,DBLP:conf/kdd/ZhouZBZWY20} incorporated knowledge graph or reinforcement learning to improve the performance of CRSs with enhanced user models or interaction mechanism.

\subsection{Dataset for Conversational Recommendation}
To facilitate the study of conversational recommendation, multiple datasets~\cite{DBLP:conf/nips/LiKSMCP18,DBLP:conf/emnlp/KangBSCBW19,DBLP:conf/acl/LiuWNWCL20,DBLP:conf/wsdm/Lei0MWHKC20} have been released in recent years.
%as shown in Table~\ref{}. [\textbf{Give a table of CRS dataset.}]
Among them, Facebook-rec~\cite{DBLP:journals/corr/DodgeGZBCMSW15} and EAR~\cite{DBLP:conf/wsdm/Lei0MWHKC20} are synthetic dialog datasets built by natural language templates based on classic recommendation datasets.
ReDial~\cite{DBLP:conf/nips/LiKSMCP18}, GoReDial~\cite{DBLP:conf/emnlp/KangBSCBW19} and DuRecDial~\cite{DBLP:conf/acl/LiuWNWCL20} are created by human annotation with pre-defined goals, such as \emph{item recommendation} and \emph{goal planning}. These goal-oriented datasets combine the elements of chit-chat and task-oriented (recommendation task) dialogs.
Compared with them, our dataset emphasizes the topic-guided process that naturally leads the conversation to the recommendation scenario in CRS.

It is worth noting that DuRecDial dataset~\cite{DBLP:conf/acl/LiuWNWCL20} is similar to TG-ReDial dataset in that it utilizes a goal sequence to guide the conversation.
However, the goal sequence is composed by multiple types of tasks (\eg recommendation, recommendation and question answering). As a comparison, TG-ReDial utilizes topic threads to characterize the evolution of content flow, which is easier to be integrated into open-domain dialogs.  
Another significant difference is that DuRecDial relies on human annotators to generate user-related data, \eg user profiles and utterances. While, TG-ReDial mainly mines suitable information from a movie review website, which closely resembles the real cases. 

\ignore{
 and the tasks are brought off by users and the bot, while the topic thread in TG-ReDial consists of different topics and is guided mainly by the bot, which is natural in multi-turn conversation.
Besides, DuRecDial produces an over-detailed user profile automatically, and annotates the conversation based on it.
Hence the user profile will greatly narrow the candidates of the discussion topics and recommended items (less than 3 items on average).
Instead, the user profile and recommended items in TG-ReDial are obtained from real data, so that it is closer to real-world scenario of CRS. 
Furthermore, we crawl the real user interaction history to help the recommendation task, which is proved essential in sequential recommendation tasks but neglected in DuRecDial.
}
\section{Dataset Construction}

\begin{figure}[tb] %图片浮动环境，类似表格中的 table [htp] 参数和表格的类似
%\vspace{-2mm}
\centering %图片居中
\includegraphics[width=\textwidth]{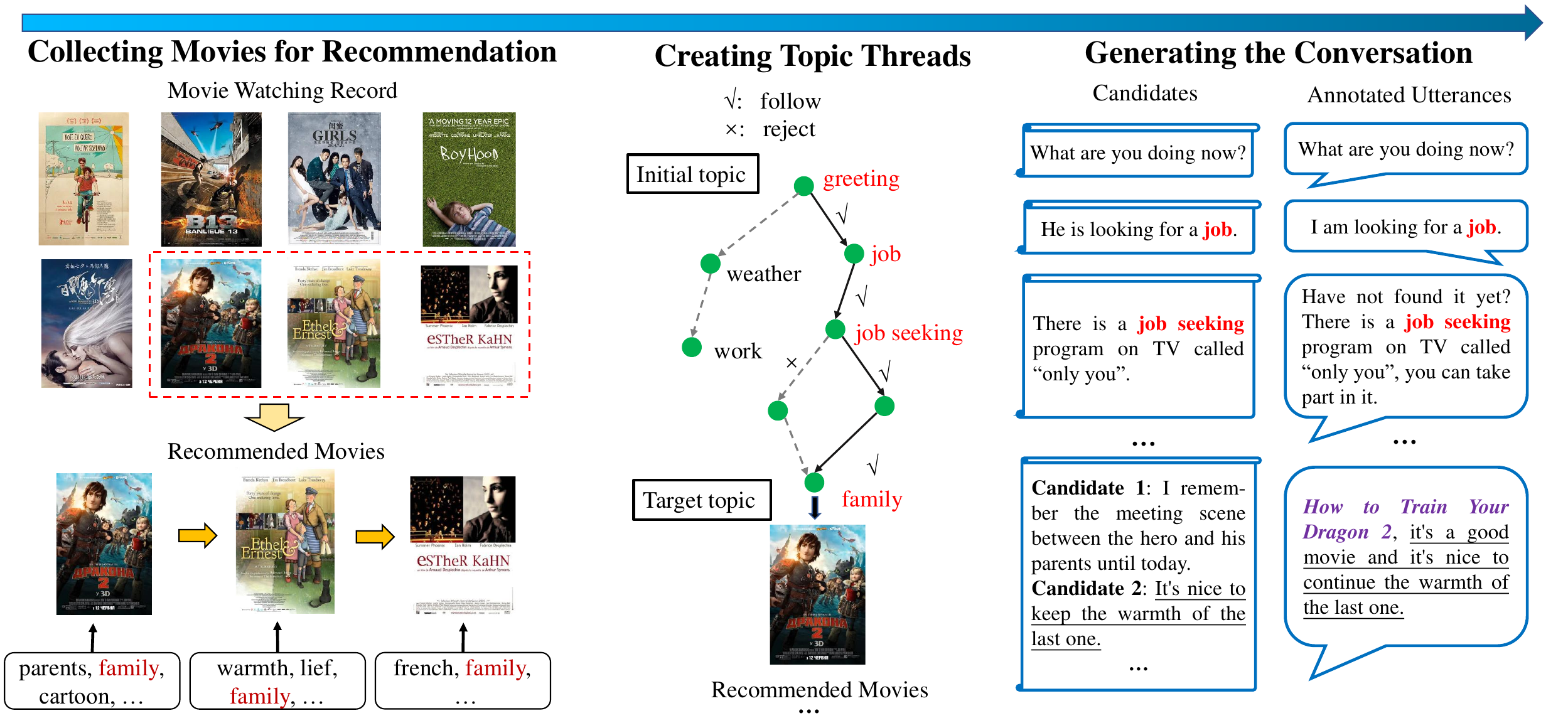}
\caption{An example of the data collection process of TG-ReDial. We select three movies sharing the same tag of ``\emph{family}'' from film watching record for recommendation,
then create the topic thread (marked in red font) on ConceptNet, and finally provide high-quality candidate sentences for human annotation.}
\label{fig-collect}
\end{figure}

%The aim of this work is to alleviate the gap between CRS and real-world application.

Following~\cite{DBLP:conf/sigir/SunZ18,DBLP:conf/nips/LiKSMCP18}, we consider a two-party setting for conversational recommendation, in which the person 
and the chat-bot play the roles of the seeker and the recommender, respectively. 
%We define one person in the dialog as the recommender (the role of CRS bots) and the other as the recommendation seeker (the role of users).
We expect that the dialogue between the seeker and the recommender starts from a non-recommendation scenario. The recommender proactively guides the conversation into the target topic, and then makes suitable recommendation considering the seeker's interests.

%, while the seeker only follows her/his preferred topics and even interrupts the guiding process of the recommender. Following~\cite{DBLP:conf/acl/TangZXLXH19,DBLP:conf/acl/WuGZWZLW19}, we define a topic to be a word (\emph{e.g.}, an entity name \emph{Emma} or a common noun \emph{book}, etc.)

%Although the above scenario is common in real world (\emph{i.e.} clothing store and cosmetics store), it is difficult to collect the conversation data.

Different from previous studies, we construct the dataset in a semi-automatic way. We utilize real data records from a popular Chinese movie review website \emph{Douban Movie}~\footnote{https://movie.douban.com/}. We associate each conversation with a real user from Douban Movie, so that the watching records (\emph{likes} and \emph{dislikes}) can be incorporated for recommendation. For a recommended movie, we create an evolving topic thread that leads from previous topic to the target topic of the movie. 
Finally, the human annotators will generate the proper reply for making recommendation based on user profiles and retrieved high-quality candidates related to the movie.
Figure~\ref{fig-collect} presents an illustrative example for the construction process.

\ignore{real movie watching records, user profiles, and review data from 
To build the conversations, we simulate this scenario based on movie viewing records from Douban Movies~\footnote{https://movie.douban.com/}.
First, we select the recommended movies in each conversation from the real data.
Then, we create the topic thread for each conversation based on knowledge graph and user profile.
Finally, we ask the annotator to complete the conversation based on retrieved high-quality candidates and above prompts.
%We leverage these movies to construct the whole dialogue between the seeker and the recommender by human annotation.
%Our data collection consists of three steps: (1) recommended movies collection; (2) topic path building; (3) conversation completing. Next we will provide details of each step. 
}

In what follows, we first describe the construction process of TG-ReDial dataset, and then present the detailed statistics about the dataset.

\subsection{Collecting Movies for Recommendation}
To simulate the real recommendation scenario, we first collect the watching records of real users on Douban for recommendation. In order to make recommendation topic-related, we attach each movie with several meaningful tags (\emph{e.g.} genre, director and starring). We keep the original tags of a movie from the Douban Movie, and further mine its reviews for extracting high-frequency keywords, and then manually select suitable tags. The number of tags for a movie is set between 1 and 38. The entire watching sequence is split into several coherent subsequence, in which the movies are ensured to share at least a common tag.
We remove incoherent subsequences. Each kept subsequence corresponds to a unique conversation, and each user is involved in four to five conversations on average. Given a user, we mark the accept/reject status about a movie according to her/his rating on it (\emph{accept}: $\ge 4$ and \emph{reject} : $\le 2$ in a five-star scale). Compared with previous studies~\cite{DBLP:conf/nips/LiKSMCP18,DBLP:conf/acl/LiuWNWCL20}, a major difference is that we reuse existing watching records with rated preferences (covering liked and disliked movies), making the generated conversations closely resemble to the real cases. 
%recommended movies based on the movie viewing record of users.

\ignore{
Each movie is equipped with several hand-picked tags including basic information from knowledge base (\emph{e.g.} producer country, director and starring) and descriptive keywords from comments (\emph{e.g.} suspense, kung fu and exciting).
Then, we split the movie viewing record from each user into several segments, to ensure that each segment owns three movies which share at least one tags. 
We define that each segment is the recommended movie set for each conversation.

Furthermore, for each user, we extract 4-5 segments to build conversation, and we leverage other movies which the user has watched before to compose the user`s historical interaction sequences.
}

\subsection{Creating Topic Threads}
\label{path_build}
Given the movies for a conversation, we incorporate topic tags to connect them in an ordered way. The initial topic of each conversation is set to \emph{greeting}, and the target topic is a selected tag of the next movie to be recommended.
For creating topic threads, we start from the initial topic and traverse over the commonsense knowledge graph ConceptNet~\cite{DBLP:conf/aaai/SpeerCH17}.
The shortest topic path identified with the \emph{Depth First Search}~(DFS) algorithm is considered as a \emph{topic thread}. We repeat the above process in multiple times until all the recommended movies can be connected via topic threads. 

\ignore{
How to guide the topics with customers (seekers) plays a key role in real-world interaction procedure for recommendation.
The initial topic of each conversation is selected from common greetings, and the target topic is a tag of the next movie waiting for recommendation.
To provide better supervision for this capability, inspired by goal-planning dataset for dialogue~\cite{DBLP:conf/acl/LiuWNWCL20,DBLP:conf/acl/WuGZWZLW19}, we build a topic thread from the initial topic to the target topic using the external commonsense knowledge graph ConceptNet~\cite{DBLP:conf/aaai/SpeerCH17}.
Specifically, we leverage DFS (Deep First Search) algorithm to find the shortest paths from the initial topic (node) to the target one.
}

In order to enhance the personality of the seeker, following~\cite{DBLP:conf/acl/KielaWZDUS18}, we generate user profiles to better control the quality of the conversation. We first collect the keywords from the profiles, self-description, and her/his review text from the original website, and then utilize 47 handwritten templates to produce sentences to describe user profiles. 
With user profiles, we can capture two options in generating topic threads, namely \emph{following} and \emph{rejection}.
At each step, the following option will incorporate the current topic in the topic thread, while the rejection option will consider another topic to extend.    
The choice is made according to whether the topic keyword appears in the extracted profile keywords. For a topic that is not among the keywords, we simply reject to extend this topic with a probability of 0.5. Such a sampling way increases the flexibility and variability of our conversation data.

\ignore{
A special phenomenon in human-to-human conversation is that the talker tends to follow topics she/he is interested in, and sometimes rejects her/his disliked one even jumps into other topics.
To construct this situation, inspired by Persona-Chat~\cite{DBLP:conf/acl/KielaWZDUS18}, we define a profile for each user (seeker).
We crawled the replies for users from Tieba~\footnote{https://tieba.baidu.com/}, and extracted top-10 high-frequency words of each user (seeker) as her/his preferred topics.
We utilize 47 handwritten templates to produce user profiles based on these topics.
Furthermore, we define two options, \emph{follow} and \emph{reject} to represent seekers` feedbacks.
}

\ignore{
Based on above preparation, we start from the initial topic (node) on ConceptNet along with the shortest path to the target one. 
On each hop, if it is one of the user preferred topics, the ``follow'' operation will be carried out and topic will be guided into the next hop. 
Once it is not the preferred topics, there is a pre-defined probability distribution for choosing the two options (We set 50\% for both.). 
For ``reject'', the seeker will provide another topic (the one the seeker prefers) to replace the current topic, and the recommender will restart to walk along the shortest path from the new topic to the target one.  
%the topic will retreat to the last hop and the recommender finds another path to the target topic. For ``reject and jump'', 
We keep the final moving path as the topic thread for annotation.
}

\subsection{Generating the Conversation}
After obtaining topic threads and recommended movies, we ask the crowd-sourced workers to complete the conversations.
Each conversation starts from chit-chat utterances, evolves according to the topic thread, and provides the  recommendation on the target topic.
Although the above information (\ie movie sequence, topic thread and user profile) has highly summarized the sketch of the conversation, it is still difficult to conduct the data annotation with a limited number of human annotators. 
%However, due to the complexity of this task, it is very hard to conduct data annotation based on these prompts.
Inspired by MultiWOZ~\cite{DBLP:conf/emnlp/BudzianowskiWTC18}, we propose a candidate-driven annotation approach based on an open-domain dialogue corpus Douban~\cite{DBLP:conf/acl/WuWXZL17} and the crawled movie reviews, to help generate the topic- and recommendation-related utterances, respectively.

%. The former is for chitchat utterances, and the latter is for movie-related utterances.

%For the two types of utterances, we leverage different annotation candidates and instructions.

%Since the topic-guided chit-chat follows the topic thread acquired in Section~\ref{path_build}, in which each utterance corresponds to one topic. 
Given a topic thread, we need generate an utterance for each topic in it. Given a topic, we first randomly retrieve 20 utterances containing the topic from Douban corpus. Then we utilize a RNN-based matching model~\cite{DBLP:conf/sigdial/LowePSP15} to compute their relevance with the last utterance, and select the most relevant sentence as the candidate utterance. In this step, the role of human annotators is to modify the retrieved candidate to ensure semantic consistency of the entire dialogue.
%Based on the high-relevance candidate, annotators just need to modify it to ensure the consistency of the whole dialogue.

\ignore{For each topic, we first randomly retrieve 20 utterances containing the topic from Douban corpus. Then we leverage a RNN-based matching model~\cite{DBLP:conf/sigdial/LowePSP15} to produce their relevance with the last utterance, and select the top-1 utterance as the candidate. Based on the high-relevance candidate, annotators just need to revise it to ensure the consistency of the whole dialogue.
}

For the target topic, we need to generate a persuasive reason for making the recommendation. Recall that the user has actually watched the movie and published a review about it. We utilize the target topic as a query to retrieve top three relevant review sentences with extreme embedding similarity~\cite{DBLP:conf/emnlp/LiuLSNCP16}. The annotators will select among the three candidate sentences, and revise or rewrite it according to conversation context when necessary.

%For recommendation, we aim to provide the recommended movie with a persuasive reason.  So we leverage comments of the recommended movie as the candidate. We select top-3 comments ranking by the relevance score, which is calculated by the extreme embedding similarity~\cite{DBLP:conf/emnlp/LiuLSNCP16} between the comment and the target topic (a tag of the movie). Annotators need to select the most persuasive one and modify it to a proper utterance.

Our crowd-sourced workers are from a specialized data annotation company. Each utterance was assigned an annotator (\emph{labeling}) and an inspector (\emph{checking}). 
Each annotator is required to carefully read the user profile and browse the detailed information on the original website.
To guarantee the quality of human-generated data, we further utilize two automatic metrics to identify low-quality cases for re-labeling. In specific, we compute the 
Distinct metrics~\cite{DBLP:conf/naacl/LiGBGD16} to filter low-informativeness dialogues with small Distinct values; and we compute the BLEU score~\cite{DBLP:conf/acl/PapineniRWZ02} between the given candidate with the human-annotated utterance, and then filter the dialogues with very little modification.
These bad cases would be relabeled until they pass the automatic evaluation.

\ignore{A issue with conditioning on conversation completing is that there is a danger that humans will, even if asked not to, unwittingly repeat utterance either verbatim or with significant word overlap.
Besides, humans tend to use the provided candidate as the utterance with little or no modification.
These problems may affect the consistency or informativeness of the conversation.
Therefore, we leverage two evaluation metrics to distinguish bad cases in annotated samples. 
We calculate the Distinct metrics for all the utterances in each dialogue and set a threshold to filter the low-informativeness dialogues which produce smaller Distinct values.
Furthermore, we compute the BLEU score between the given candidate with the human-annotated utterance and filter the dialogues with high BLEU values.
These bad cases are relabeled until pass the thresholds.
}

\subsection{The TG-ReDial Dataset}

\begin{table}[t]
\begin{center}
    \small
	\begin{tabular}{l|r|l|r}
	\hline
	 & Total & & Average \\
	\hline
	\hline
	\# Users & 1,482 &\# Words per Utterance &  19.0 \\
	\# Dialogues & 10,000 &\# Topics per Dialogue & 7.9 \\
	\# Utterances & 129,392 &\# Movies per Dialogue & 3 \\
	\# Movies & 33,834 &\# Profile Sentences per User & 10\\
	\# Topics & 2,571 &\# Watched Movies per User & 202.7\\
	%\# Avg. Words per Utterance &  19.0\\
	%\# Avg. Topics per Dialogue & 7.9 \\
	%\# Avg. Accepted/Rejected Topics per Dialogue & 7.4/0.6\\
	%\# Avg. Movies per Dialogue & 3 \\
	%\# Avg. Profile Sentences per User & 10 \\
	%\# Avg. Interacted Movies per User & 202.7 \\
	\hline
	\end{tabular}
	%}
	\caption{Data statistics of our TG-ReDial dataset.}
\label{statics}
\end{center}
\end{table}

%\paragraph{Corpus Statistics}
The detailed statistics of TG-ReDial are shown in Table~\ref{statics}.
TG-ReDial consists of 129,392 utterances from 1,482 users.
%, it is sufficient to train deep models or fine-tune the pre-trained models.
Our dataset is constructed in a topic-guided way, containing more informative sentences. 
On average, a dialogue has 7.9 topics and an utterance contains 19.0 words, which are larger than the corresponding numbers of existing CRS datasets~\cite{DBLP:conf/nips/LiKSMCP18,DBLP:conf/acl/LiuWNWCL20,DBLP:conf/emnlp/KangBSCBW19}.
% It indicates the informativeness of our dataset.
Furthermore, a user has 10 profile sentences and 202.7 watching records on average. 
%They are produced based on real data and provide rich background information of users.

A major feature of our dataset is that we organize the conversation by topic threads, so that the transitions from chit-chat to  recommendation are more natural. 
Such a dataset is particularly useful to help integrate the recommender component into general-purpose chat-bots, since it is easy to align our topics with open-domain conversations. 
Moreover, we associate  a conversation with a unique user identity, so that it can closely resemble real-world cases. Especially, we can obtain profiles and watching history for the users in a conversation. 
To our knowledge, most of existing datasets~\cite{DBLP:conf/nips/LiKSMCP18,DBLP:conf/acl/LiuWNWCL20} mainly focus on cold-start scenario for CRS, while it is also important  that CRS can leverage historical interaction data for existing  users. Our dataset provides the possibility of training conversational recommendation algorithms with historical interaction data. It is also feasible to study  other personalized tasks, since a user is involved in multiple conversations in our dataset.

Note that, in order to protect user privacy, we only sample users with a large number of watching records. For derived user data (\eg profile or watching records), we perform the anonymized operation and add randomized modification (\eg removal, replacement or deletion). We also require that the retrieved review sentences have to be written via paraphrasing.
Finally, we ask human annotators to manually trace the user identities with corresponding user data in our dataset.   We do not include the data from the users that can be identified in the final dataset. 

%Based on the user interaction history, some advanced models in sequential recommendation~\cite{DBLP:conf/icdm/KangM18,DBLP:conf/icdm/LiuWWLW16} can be used to improve the recommendation performance. Besides, the user preferred topics are helpful to generate more satisfying responses to users.
%Compared with conventional CRS datasets~\cite{} which randomly select recommended item(movie) set from all items(movies), our approach leverages the segments from real movie records of users, so that movies recommended in each conversation share implicit correlation.

%Compared with other CRS dataset which build the dialogue by special games or just crawl from websites, our pre-defined topic path can ensure that the topic is gradually transferred into the target topic in multi-turn conversation, and the whole process is natural and purposeful.

\section{Our Approach}
%We aim at developing a conversational recommender system, which is able to chat with a user in the purpose of topic-guiding, and make persuasive movie recommendations.
%Following classic CRS~\cite{}, we propose a framework consisting of a recommendation module and a dialog module.
In this section, we first formulate the topic-guided conversational recommendation task. Then we introduce our solution to this task.
% consisting of a recommendation module and a dialog module.

\subsection{Problem Formulation}
\label{problem}
Given a user $u$, we assume that she/he is associated with a profile $P_{u}$ (a set of descriptive sentences related to the topics that $u$ is interested in) and a historical interaction sequence $I_{u}$ (a chronologically-ordered sequence of items that $u$ has interacted with). Each dialogue is composed by a list of utterances, denoted by $d=\{s_{k}\}^{n}_{k=1}$, in which $s_{k}$ is the utterance at the $k$-th turn. We consider the CRS in a topic-guided manner, and each utterance $s_{k}$ is associated with a topic $t_k$. When $t_k$ is a target topic, the system will trigger the recommendation of item $i_k$ with the persuasive reason.

%means that $s_{k}$ will include a recommended item $i_k$ with the persuasive reason.  

%Let  $D^{u}=\{d_{1},\cdots,d_{m}\}$ denote a set of dialogues for each user $u$, where $m$ is the number of dialogues for user $u$.

\ignore{

Assume that we have a set of users and items, denoted by $\mathcal{U}$ and $\mathcal{I}$.
For each user $u$, we attach a user profile $P^{u}$ and an interaction sequence $I^{u}$ as her/his background information.
The user profile $P_{u}$ consists of 10 sentences, each sentence gives a description of a user preferred topic.
$I^{u}$ is a chronologically-ordered interaction sequence with items: $\{i_{1},\dots,i_{n}\}$, where $n$ is the number of interactions and $i_{k}$ is the $k$-th item that the user $u$ has interacted with.
%It is worth noting that the user profile $P^{u}$ and the interaction sequence $I^{u}$ are very helpful to dialog and recommendation module, respectively.
There are a set of dialogues $D^{u}=\{d_{1},\cdots,d_{m}\}$ for each user $u$, where $m$ is the number of dialogues for user $u$.
Each dialogue is composed by a list of utterances, denoted by $d=\{s_{k}\}^{n}_{k=1}$, in which each utterance $s_{k}$ is a conversation sentence at the $k$-th turn.
Recall that there are two stages for topic-guided chit-chat and persuasive recommendation in each dialogue $d$, respectively. 
Hence each utterance $s_{k}$ for the recommender from topic-guided chit-chat or persuasive recommendation is associated with a specific topic $t_{k}$ or recommended movie $i_{k}$, respectively.
}

Based on these notations, the task of topic-guided conversational recommendation is defined as: given the user profile $P_{u}$, user interaction sequence $I_{u}$, historical utterances $\{s_{1},\dots,s_{k-1}\}$ and corresponding topic sequence $\{t_{1},\dots,t_{k-1}\}$, we aim to (1) predict the next topic $t_{k}$ to reach the target topic, or (2) recommend  the movie $i_{k}$, and finally (3) produce a proper response $s_{k}$ about the topic or with persuasive reason.
The three sub-tasks are referred to topic prediction, item recommendation and response generation. 

\subsection{Recommendation Module}
The recommendation module aims to predict the item that a user likes given the conversation context.
The key point is how to derive an effective user presentation for recommendation.
We consider two kinds of data signals for this task. 
Specially, we utilize the pre-trained language model BERT~\cite{DBLP:conf/naacl/DevlinCLT19} to encode the historical utterances $\{s_{1},\dots,s_{k-1}\}$, and a self-attentive sequential recommendation model SASRec~\cite{DBLP:conf/icdm/KangM18} to encode the user interaction sequence $I_{u}$.

The representation $\bm{v}_u$ of user $u$ is obtained as follows:
\begin{eqnarray}
\label{eq-user}
\bm{v}_u=\text{MLP}([\bm{v}_u^{(1)};\bm{v}_u^{(2)}]),
\end{eqnarray}
where $\bm{v}_u^{(1)}$ (obtained from BERT) and $\bm{v}_u^{(2)}$ (obtained from SASRec) are the embeddings to represent the historical utterances and interaction sequence, respectively.
%is the representation of historical utterances, it is the output of BERT in the first position ([\texttt{CLS}]).$\textbf{v}^{(i)}$ denotes the representation of the user interaction sequence, it is the learned state of the last position from SASRec.
Given the user representation, we can compute the probability that recommends an item $i$ from the item set to a user $u$:
\begin{eqnarray}
\label{eq-rec}
\text{P}_{rec}(i)=\text{softmax}(\bm{e}_{i}^{\top} \cdot \bm{v}_{u})
\end{eqnarray}
where $\bm{e}_{i}$ is the learned item embedding for item $i$. We utilize Equation~\ref{eq-rec} to rank all the items and select the item with the largest probability for recommendation.

\subsection{Dialog Module}
The dialog module aims to generate proper responses to the user (seeker) for topic guidance or item recommendation. We achieve the two purposes with specific models. %For achieving the two purpose, we propose two  
%For guiding the topic accurately, we divide the dialog module into two sub-model, a topic prediction model and a response generation model.

\paratitle{Topic Prediction Model.} It predicts the next topic that guides user $u$ towards the target topic. We mainly utilize text data for topic prediction, and implement three different BERT-based encoders, namely conversation-BERT, topic-BERT and profile-BERT, for encoding the historical utterances, historical topic sequence and user profile, respectively.
For each BERT variant, we simply concatenate all the available text data and the target topic (with [\texttt{SEP}] tokens for separating). The incorporation of target topic is to enhance the topical semantics. 
Based on the obtained representations, we compute the probability of a topic $t$ as the next topic by:
\begin{eqnarray}
\label{eq-topic}
\text{P}_{topic}(t)=\text{softmax}(\bm{e}_{t}^{\top} \cdot \text{MLP}([\bm{r}^{(1)};\bm{r}^{(2)};\bm{r}^{(3)}]),
\end{eqnarray}
where $\bm{e}_{t}$ is the learned embedding for topic $t$, $\bm{r}^{(1)}$, $\bm{r}^{(2)}$ and $\bm{r}^{(3)}$ are the embeddings of historical utterances, topics and user profile obtained from conversation-BERT, topic-BERT and profile-BERT, respectively. We utilize Eq.~\ref{eq-topic} to rank all the topics and select the topic with the largest probability.

\paratitle{Response Generation Model.}
It aims to generate proper responses for topic-guided conversations.
We leverage the pre-trained text generation model GPT-2~\cite{radford2019language} for response generation. GPT-2 utilizes a stacked of masked multi-head self-attention layers trained on massive web-text data by the generic language model~\cite{radford2019language,DBLP:conf/naacl/DevlinCLT19}.
We consider two cases in this model. 
For non-recommendation case, we generate the response conditioned on the predicted topic, and concatenate $t_{k}$ with the historical utterances $\{s_{1},\dots,s_{k-1}\}$ (separated by [\texttt{SEP}] tokens). For recommendation case, we generate the persuasive reason conditioned on the recommended item, and concatenate the recommended movie $i_{k}$ with the historical utterances $\{s_{1},\dots,s_{k-1}\}$.
For both cases, we can unify the input as a long sequence, which will be encoded and fed into GPT-2 for decoding.

\section{Experiments}
%In this section, we first set up the experiments, and then report the results and analysis.
%We split TG-CRS into train/dev/test data by randomly sampling 85\%/7.5\%7.5\% data. 
We evaluate the proposed approach on TG-ReDial dataset, which is split into training, validation and test sets using a ratio of 8:1:1.
%at the level of users to avoid data leakage. 
For each conversation, we start from the first utterance, and  generate reply utterances or recommendations in turn by our model.
We perform the evaluation on the three sub-tasks, namely item recommendation, topic prediction  and response generation.
%In TG-CRS task, we consider three sub-tasks for evaluation, recommendation task, topic prediction task and response generation task.
%We compare the performance of our model with several competitive models in the three sub-tasks.

\subsection{Evaluation on Item Recommendation}
In this subsection, we conduct a series of experiments on the effectiveness of our proposed model for the recommendation task.
Following~\cite{DBLP:conf/icdm/KangM18,DBLP:conf/icdm/LiuWWLW16}, we adopt NDCG@$k$ and MRR@$k~(k=10,50)$ as evaluation metrics for ranking all the possible items. 
%of the ground-truth items in all possible items for evaluation.

\paragraph{Baselines.}
We consider the following baselines for performance comparisons:
(1) \emph{Popularity} ranks items according to popularity measured by the number of interactions.
(2) \emph{ReDial}~\cite{DBLP:conf/nips/LiKSMCP18} is proposed specially for CRSs by utilizing an auto-encoder for recommendation.
(3) \emph{KBRD}~\cite{DBLP:conf/emnlp/ChenLZDCYT19} is the state-of-the-art CRS model using knowledge graphs to enhance the semantics of contextual items or entities for recommendation.
(4) \emph{GRU4Rec}~\cite{DBLP:conf/icdm/LiuWWLW16} applies GRU to model user interaction history without using conversation data. %We represent the items using embedding vectors rather than one-hot vectors.
(5) \emph{SASRec}~\cite{DBLP:conf/icdm/KangM18} adopts the Transformer architecture to encode user interaction history without using conversation data. 
(6) \emph{TextCNN}~\cite{DBLP:conf/emnlp/Kim14} adopts a CNN-based model to extract textual features from contextual utterances for recommendation.
(7) \emph{BERT}~\cite{DBLP:conf/naacl/DevlinCLT19} is a pre-training language model that directly encodes the concatenated  historical utterances.

%Among all the methods, Popularity is a heuristic method. ReDial and KBRD are recently proposed CRS models, but they only focus on modeling the movies present in historical utterances.
%GRU4Rec and SASRec are sequential recommendation models, which only utilize the user interaction sequence.
%TextCNN and BERT are text-based models, which only utilize the historical utterances.
%Our proposed model is the mixture of SASRec and BERT, it can utilize the user interaction sequence and historical utterances jointly.

\begin{table}[t]
\begin{center}
    \small
	\label{tab:datasets}
	\begin{tabular}{l||c|c|c|c}
    \hline
     Models &NDCG@10 &NDCG@50 &MRR@10 &MRR@50\\
    \hline
    \hline
    Popularity &0.0015 &0.0036 &0.0011 &0.0015 \\
    ReDial & 0.0009 &0.0029 &0.0005 &0.0009\\
    KBRD & 0.0064& 0.0111 & 0.0040 & 0.0049 \\
    GRU4Rec & 0.0028 &0.0062 & 0.0014 &0.0020 \\
    SASRec & 0.0092& 0.0179& 0.0050 & 0.0068 \\
    TextCNN & 0.0144& 0.0215& 0.0119 &0.0133 \\
    BERT & 0.0246& 0.0439& 0.0182 & 0.0221 \\
    \hline
    \textbf{Ours} & \textbf{0.0348}& \textbf{0.0527}& \textbf{0.0240} &\textbf{0.0277}\\
    \hline
  \end{tabular}
	%}
	\caption{Results on item recommendation task. }
	%Numbers marked with * indicate that the improvement is statistically significant compared with the best baseline (t-test with p-value $< 0.05$).}
\label{rec-table}
\end{center}
\end{table}

\paragraph{Result and Analysis.} Table~\ref{rec-table} presents the performance of different methods on recommendation task.
As we can see, 
%there is a performance order that ours \textgreater text-based model \textgreater sequential recommendation model \textgreater CRS models. 
%It indicates that the historical utterances own 
Popularity performs better than ReDial but worse than KBRD.
ReDial and KBRD utilize the items in historical utterances for recommendation. Besides,  KBRD incorporates external knowledge graph to enhance the representations of items.
Second, SASRec outperforms GRU4Rec and the two CRS models (\ie KBRD and ReDial). 
%Sequential models utilize the user interaction history to represent the user preference, which contains many of items.
It indicates that self-attentive architecture is particularly suitable for modeling the interaction history.
Furthermore, text-based TextCNN  (\ie TextCNN and BERT) perform better than other baselines, which indicates that it is useful to leverage the historical utterances for recommendation.
Among the two text-based models, BERT outperforms TextCNN, since it is adopts more powerful architecture trained with large-scale data.
Finally, our proposed model outperforms all the baselines significantly. Our model is able to utilize both historical utterances and interaction sequence, combing the merits of BERT and SASRec. 

%Since our model leverages BERT and SASRec to encode the historical utterances and interaction sequence jointly.

\subsection{Evaluation on Topic Prediction}
We continue to evaluate the performance of our approach on the topic prediction task. Following~\cite{DBLP:conf/acl/TangZXLXH19}, we adopt Hit@$k~(k=1,3,5)$ as evaluation metrics for ranking all the possible topics. 

\paragraph{Baselines.}
We consider the following baselines for performance comparison:
%(1) \textbf{Embedding} is a heuristic method that ranks topics according to the cosine similarity with the last topic.
(1) \emph{PMI} measures the point-wise mutual information with the last topic for ranking.
(2) \emph{MGCG}~\cite{DBLP:conf/acl/LiuWNWCL20} is a recently proposed CRS model based on multi-type GRUs (with a special GRU to encode user profiles).
(3) \emph{Conversation/Topic/Profile-BERT} utilizes the conversation/topic/profile-BERT to encode historical utterances/topics/user profiles for predicting the next topic, respectively.
%(4) \textbf{Ours w/o conversation-BERT/topic-BERT/profile-BERT} is the ablation models of our proposed model. We remove the conversation-BERT, topic-BERT or profile-BERT, respectively.
(4) \emph{Ours w/o target} is the ablation model of our proposed model by removing the target topic from input.

%Among all the methods, PMI is a heuristic method. LSTM considers the three types of information with widely-used LSTM models. Conversation-BERT/topic-BERT/profile-BERT are 
%Ours w/o conversation-BERT/topic-BERT/profile-BERT and 
%Ours w/o target topic is the ablation models of our proposed models, to validate the effective of these parts in our model.

\begin{table}[t]
\begin{center}
    \small
	\begin{tabular}{l||c|c|c}
    \hline
     Models &Hit@1 &Hit@3 &Hit@5\\
    \hline
    \hline
    PMI &0.0349  &0.0927 &0.1290\\
    MGCG &0.6098  &0.8128 &0.8294\\
    Conversation-BERT &0.6114  &0.8189 &0.8341\\
    Topic-BERT &0.6155  &0.8275 &0.8405\\
    Profile-BERT &0.4986  &0.8205 &0.8344 \\
    \hline
    Ours \emph{w/o} target &0.4420 &0.5923 &0.6374\\
    \textbf{Ours} & \textbf{0.6231}& \textbf{0.8370} &\textbf{0.8497}\\
    %\hline
    %Ours w/o conversation-BERT & & & \\
    %Ours w/o topic-BERT &  & &\\
    %Ours w/o profile-BERT & 0.6619 & 0.6999 &\\
    \hline
  \end{tabular}
	%}
	\caption{Results on topic prediction task.} 
	%Numbers marked with * indicate that the improvement is statistically significant compared with the best baseline (t-test with p-value $< 0.05$).}
\label{topic-table}
\end{center}
\end{table}

\paragraph{Result and Analysis.} Table~\ref{topic-table} presents the performance of different methods on  topic prediction task.
As we can see, PMI does not perform well, since it cannot consider the target topic.
Second, Conversation/Topic-BERT performs better than MGCG. It indicates that the pre-trained language model BERT is particularly  suitable to capture topical semantics.
Among the BERT-based models, Profile-BERT performs worse, since historical utterances and topics are more important to consider in this task.
Furthermore, our model outperforms all the baselines, since it jointly utilizes historical utterances, topics and user profiles encoded by different BERT models.
Finally, after removing target topic from the input of BERT models, the performance decreases significantly, indicating that the target topic is important.

\subsection{Evaluation on Response Generation}
Finally, we evaluate the performance of our approach on the response generation task. Following~\cite{DBLP:conf/emnlp/ChenLZDCYT19,DBLP:conf/acl/QiuLBZY19,DBLP:conf/ijcai/TaoGSWZY18}, we adopt perplexity (PPL) and BLEU-1,2,3 for evaluating the relevance between generated response with the ground truth, and adopt Distinct-1,2 for evaluating the informativeness of the generated utterances. Furthermore, we invite human annotators to score the \emph{Relevance}, \emph{Fluency} and \emph{Informativeness} of the generated results with the rating range of $[0,2]$.

\paragraph{Baselines.}
We consider the following baselines for performance comparison:
(1) \emph{ReDial}~\cite{DBLP:conf/nips/LiKSMCP18} adopts the hierarchical RNN for response generation.
(2) \emph{KBRD}~\cite{DBLP:conf/emnlp/ChenLZDCYT19} applies Transformer with enhanced modeling of word weight based on knowledge graphs.
(3) \emph{Transformer}~\cite{DBLP:conf/nips/VaswaniSPUJGKP17} applies a Transformer-based encoder-decoder framework to generate proper responses.
(4) \emph{GPT-2}~\cite{radford2019language} is a pre-training text generation model and fine-tuned on TG-ReDial dataset.

%(5) \textbf{GPT-2+NS}~\cite{DBLP:conf/iclr/HoltzmanBDFC20} adopts the nucleus sampling based on GPT-2.

%Among all the methods, ReDial and KBRD are recent CRS models. Transformer is a popular text generation model. GPT-2 is pre-trained text generation model.
%Compared with above models, our proposed model appends the predicted topic/item into input sentence.

\begin{table}[t]
\begin{center}
    \small
	\begin{tabular}{l||c|c|c|c|c|c}
    \hline
     Models &PPL &BLEU-1/2/3 &Distinct-1/2 &Relevance &Fluency &Informativeness \\
    \hline
    \hline
    ReDial &81.614 &0.177/0.028/0.006 &0.004/0.025 &0.660 &0.400 &0.560 \\
    KBRD &28.022  &0.221/0.028/0.009 &0.004/0.008 & 0.860 &1.060 &1.300\\
    Transformer &32.856  &\textbf{0.287}/\textbf{0.071}/\textbf{0.032} &0.013/0.083 &1.400 &1.360 &1.440 \\
    GPT-2 &13.383  & 0.279/0.066/\textbf{0.032}
    &0.017/\textbf{0.094}  &1.460 &\textbf{1.440} &1.500 \\
    \hline
    \textbf{Ours} & \textbf{7.223} & 0.280/0.065/0.031 & \textbf{0.021}/\textbf{0.094}
    & \textbf{1.520} &1.360 &\textbf{1.640} \\
    \hline
  \end{tabular}
	%}
	\caption{Results on response generation task.} 
	%Numbers marked with * indicate that the improvement is statistically significant compared with the best baseline (t-test with p-value $< 0.05$).
\label{gene-table}
\end{center}
\end{table}

\paragraph{Result and Analysis.} Table~\ref{gene-table} presents the performance of different methods on response generation task.
The first observation is that ReDial does not perform well on our dataset. A major reason is that ReDial utilizes a hierarchical RNN for response generation, which is not suitable to encode long utterances (recall that utterances in our dataset are longer and more informative).
Second, Transformer is better than KBRD in most of metrics, since KBRD  utilizes knowledge graph information to promote the predictive probability of entities and items, which may have an adverse effect on text generation.
Furthermore, Transformer, GPT-2 and our model give similar BLEU scores. For PPL, Distinct and human evaluation, Transformer achieves the worst results, while our model achieves very good results. Indeed, BLEU may not be suitable for evaluating CRSs~\cite{DBLP:conf/aaai/TaoMZY18}, because it is easier to be affected by meaningless words such as stopwords.
%It indicates that BLEU is not suitable to evaluate the generated responses in CRS, because it is easily affected by meaningless stopwords~\cite{DBLP:conf/aaai/TaoMZY18}.
%GPT-2 and ours leverage pre-trained Transformer model so that provide more fluent and diverse results.
Finally, our model outperforms all the baselines in most cases, since it can utilize the predicted topic or item to enhance the quality of the generated text.

% Since our model leverages the predicted topics/items to control the generation process.
%But in Fluency, GPT-2 performs better than ours, the reason may be that the given topic may hurt the performance of the language model.
\section{Conclusion}
We introduced  a high-quality dataset TG-ReDial for conversational recommender systems, which was constructed by human annotation based on  real-world user data. Based on TG-ReDial, we presented the task of topic-guided conversational recommendation and a solution to this task.
Extensive experiments have demonstrated the effectiveness of the proposed approach on three sub-tasks. 
%By constructing extensive experiments, our framework yielded better performance than several competitive baselines.

\ignore{
Compared with other CRS datasets, TG-ReDial focused on the topic guiding process and persuasive recommendation in CRS, and provided rich background information for users.
Based on TG-ReDial, we presented a novel task, topic-guided conversational recommender system (TG-CRS), consisting of three sub-tasks.
Then we proposed a targeted framework to handle it, which contains a recommendation and a dialog module.
By constructing extensive experiments, our framework yielded better performance than several competitive baselines.
}

Currently, the potential of TG-ReDial dataset has not been fully explored. It can be useful as a testbed for more tasks, such as personalized chit-chat~\cite{DBLP:conf/acl/KielaWZDUS18}, target-guided conversation~\cite{DBLP:conf/acl/TangZXLXH19} and sequential recommendation~\cite{DBLP:conf/cikm/ZhouWZZWZWW20}.
 As future work, we will investigate the study of these tasks on TG-ReDial dataset. Besides, we will also consider how to construct more effective approaches to topic-guided conversational recommendation. 

\section*{Acknowledgement}
This work was partially supported by the National Natural Science Foundation of China under Grant No. 61872369 and 61832017,  Beijing Academy of Artificial Intelligence (BAAI) under Grant No. BAAI2020ZJ0301, and Beijing Outstanding Young Scientist Program under Grant No. BJJWZYJH012019100020098, the Fundamental Research Funds for the Central Universities, the Research Funds of Renmin University of China under Grant No.18XNLG22 and 19XNQ047. Xin Zhao is the corresponding author.

% include your own bib file like this:
\bibliographystyle{coling}
\bibliography{coling2020}

\end{document}